\documentclass[11pt]{article}
\usepackage{libertine}
\usepackage[margin=1in]{geometry}
\usepackage{amsmath,amssymb}
\usepackage{graphicx}
\usepackage[authoryear]{natbib}
\usepackage{hyperref}
\usepackage[latin9]{inputenc}
\usepackage{babel}
\begin{document}
\title{Guided Discrete Diffusion for Constraint Satisfaction Problems}
\author{Justin Jung}
\date{Jan 13, 2025} 
\maketitle

\subsection*{Introduction}

AI for constraint satisfaction problems is an important field researched
for more than half a century. Sudoku, a puzzle where no row, column,
or block can have two of the same number, is a popular benchmark to
assess the ability of models to reason over constraints. With the
rise of deep learning, many deep neural networks have been used to
solve sudoku, such as transformers and graph neural networks to name
a few. These networks perform well but are trained under supervision
and assume access to a labelled dataset. Given the importance of identifying
patterns in the structure of Sudoku solutions, these supervised methods
may be limited in their capability to generalize to unseen puzzles
(as combinatorially many exist) or perform well under limited data
settings---and at the minimum, require a supervised dataset of initial
puzzles $x$ to final puzzle solutions $y$.

Thus we instead employ an unsupervised generative modelling approach
to learn the distribution of sudoku puzzles in hopes that our model
learns moreso the structures and patterns of sudoku puzzles. Diffusion
is a widely used generative model for all kinds of settings, and here
we apply it to learn Sudoku puzzles. Since sudoku (a board that is
a nine by nine matrix filled with numbers one through nine) is inherently
a matrix of discrete numbers, we use a discrete diffusion model. While
continous diffusion (such as the popular DDPM) can be used if sudoku
boards are relaxed and represented in continuous space $\mathbb{R}^{9\times9}$,
we believe that preserving the discrete nature of Sudoku puzzles is
more natural and so opt for a discrete diffusion model. 

\subsection*{Background}

Diffusion at a very high level is a generative model defined by a
forward markov chain (which ``corrupts'' the actual data distribution
we care about to some prior distribution easy to sample from, such
as complete noise); it aims to learn the reverse markov chain which
can take a sample from the prior and progressively ``decorrupt it''
and transform it to a sample who comes from a distribution approximate
to the actual data distribution we wish to learn. (For a more extensive
treatment of diffusion models, Calvin Luo has a good tutorial \cite{luo2022understanding}). 

For discrete data, \cite{austin2021structured} introduce Discrete Denoising
Diffusion Probabilistic Models. (For a more extensive treatment, please
refer to their work ``Structured Denoising Diffusion Models in Discrete
State-Spaces''). A sequence $\mathbf{x}_{t}\in\mathbb{Z}^{L}$ of
$L$ many discrete categorical variables of $K$ categories $x\in[K]$
has a forward corrupting markov chain defined by a transition kernel
matrix $Q_{t}\in\mathbb{R^{K\times K}}$. In the forward process,
the transition kernel matrix $Q_{t}$ is applied to all tokens or
categorical variables in the sequence $x_{t}$ independently (categorical
variables are represented as one-hot vectors $x\in\mathbb{Z}^{K}$);
thus for clarity we can just focus on the case where our sequence
has one token $L=1$ and we collapse to a single categorical variable
$x$. The one timestep forward transition probability of a categorical
variable $x$ is the categorical distribution induced by applying
the kernel matrix to the previous state of $x$; $q(x_{t}|x_{t-1})=Cat(x_{t};p=x_{t-1}Q_{t})$
. Discrete Denoising Diffusion Probabilistic Models (D3PM) only work
with discrete time markov chains; however, as seen later, more general
continuous time diffusion models exist as well. 

\begin{figure}

\begin{centering}
\includegraphics[scale=0.3]{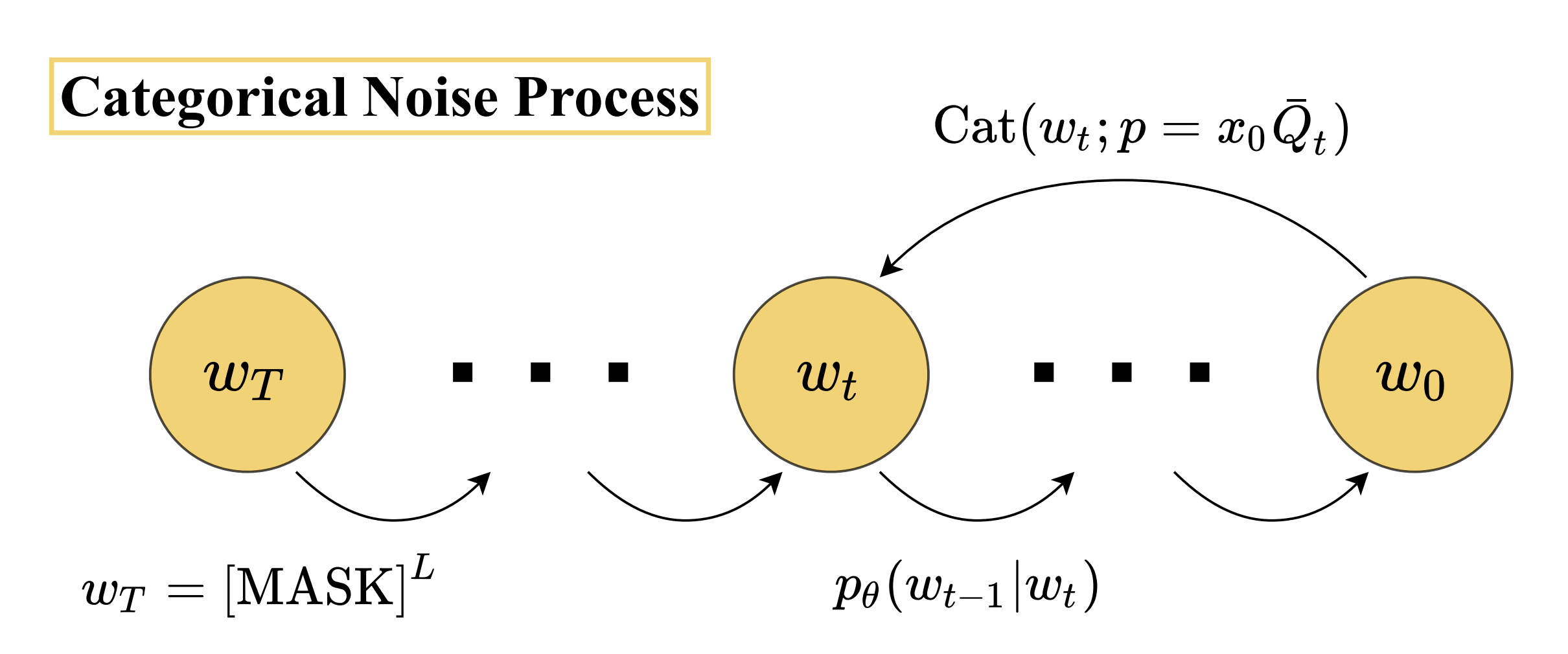}\caption{Categorical Noise Process \cite{gruver2023protein}}
\par\end{centering}
\end{figure}

In the discrete time setting, the goal of diffusion is to learn the
reverse transition process at each timestep, which is defined by $p_{\theta}(x_{t-1}|x_{t})$.
While this one step reverse transition can be directly modelled by
a network, \cite{austin2021structured} parameterize the one step transition through
a denoiser $\tilde{p}_{\theta}(\tilde{x_{0}}|x_{t})$: 
\[
p_{\theta}(x_{t-1}|x_{t})\propto\sum_{\tilde{x_{0}}}q(x_{t-1},x_{t}|\tilde{x_{0}})\tilde{p_{\theta}}(\tilde{x_{0}}|x_{t})
\]

where the denoiser is learned with a neural network. In brief this
parametrization comes from applications of Bayes rule, where $q(x_{t-1}|x_{t},\tilde{x_{0}})=\frac{q(x_{t}|x_{t-1},x_{0})q(x_{t-1}|x_{0})}{q(x_{t}|x_{0})}$
is derived using Bayes and we also have 
\[
q(x_{t-1},x_{t}|\tilde{x_{0}})\tilde{p_{\theta}}(\tilde{x_{0}}|x_{t})=q(x_{t-1}|x_{t},\tilde{x_{0}})q(x_{t}|\tilde{x_{0}})\tilde{p_{\theta}}(\tilde{x_{0}}|x_{t})\propto q(x_{t-1}|x_{t,}\tilde{x_{0}})\tilde{p_{\theta}}(\tilde{x_{0}}|x_{t})
\]

where with our construction of the forward process we have $q(x_{t}|\tilde{x}_{0})=Cat(x_{t};p=x_{0}\bar{Q}_{t})$
where $\bar{Q}_{t}=\prod_{i=1}^{t}Q_{i}$ is the cumulative product
of the absorbing transition kernels over time and we can also calculate
(derivation is slightly involved) $q(x_{t-1}|x_{t},x_{0})$ . 

One thing that is important to note is that in the forward process,
our transition kernel matrix $Q_{t}$ is applied to each token in
the sequence independently, meaning that each categorical variable
forward diffuses independent of the other tokens. However, in the
reverse diffusion process, this is not the case---our denoiser network
$\tilde{p}_{\theta}(\tilde{x_{0}}|x_{t})$ is a neural network that
conditions on all tokens in the sequence $\mathbf{x}_{t}=[x_{t}]_{l=1}^{L}.$

This parameterization of a denoiser $\tilde{p}_{\theta}(\tilde{x_{0}}|x_{t})$
allows for flexible calculation of transitions, such as skipping k-steps
at once 
\[
p_{\theta}(x_{t-k}|x_{t})\propto\sum_{\tilde{x_{0}}}q(x_{t-k},x_{t}|\tilde{x_{0}})\tilde{p_{\theta}}(\tilde{x_{0}}|x_{t})
\]

\cite{austin2021structured} consider various transition matrices $Q_{t}$; we choose
the simple absorbing transition kernel where a state transitions to
absorbing state $[MASK]$ with probability $\beta_{t}$, else stays
the same. This simple transition kernel matrix allows for easy calculation
of $q(x_{t}|x_{0})=Cat(x_{t};p=x_{0}\bar{Q}_{t})$. 

For the general transition kernel, we have as our diffusion loss objective
for our denoiser $\tilde{p}_{\theta}(\tilde{x_{0}}|x_{t})$ as 
\[
L_{\lambda}=L_{VB}+\lambda\mathbb{E}_{q(x_{0})}\mathbb{E}_{q(x_{t}|x_{0})}[-log\tilde{p}_{\theta}(x_{0}|x_{t})]
\]
 where the variational lower bound loss is defined as 
\[
L_{VB}=\mathbb{E}_{q(x_{0})}\big[D_{KL}[q(x_{T}|x_{0})||p(x_{T})]+\sum_{t=2}^{T}\mathbb{E}_{q(x_{t}|x_{0})}\big[D_{KL}[q(x_{t-1}|x_{t},x_{0})||p_{\theta}(x_{t-1}|x_{t})]\big]-\mathbb{E}_{q(x_{1}|x_{0})}\big[logp_{\theta}(x_{0}|x_{1})\big]\big]
\]

Fortunately, \cite{austin2021structured}. conveniently show that the generative masked
language model (MLM) objective (inferring the unmasked token of a
masked token, as in the LLM BERT objective) is equivalent to our diffusion
loss objective $L_{\lambda}$ under the simple absorbing transition
kernel. Thus it is enough for our model to infer the unmasked token
of a masked token---and consequently it is enough for our loss function
to simply be the cross entropy or log loss over the masked tokens. 

Thus, we define the simple likelihood objective of predicting the
unmasked tokens. With abuse of notation, we can simply write this
as the cross entropy loss of predicting the actual unmasked sequences
$w_{0}$ drawn from our actual data distribution $q(w_{0})$ given
the masked sequences $w_{t}$ drawn from the forward transition distribution
$p(w_{t}|w_{0})$:
\[
L(\theta)=\mathbb{E}_{w_{0},t}\big[-logp_{\theta}(w_{0}|w_{t})\big]\qquad w_{t}\sim p(w_{t}|w_{0})
\]

More accurately, our cross entropy loss is calculated only over the
tokens which are masked (say denoted by set of indices $I$)
\[
L(\theta)=\mathbb{E}_{w_{0},t}\big[\frac{1}{I}\sum_{i:[w_{t}]_{i}=[MASK]}-logp_{\theta}([w_{0}]_{i}|w_{t})\big]\qquad w_{t}\sim p(w_{t}|w_{0})
\]

As usual we assume that our given dataset $\{w_{0}\}$ is drawn i.i.d
from the actual data distribution $q(x_{0})$. To implement this loss
in practice, our loss function is calculated over minibatches defined
from the given training data $\{w_{0}\}$; for each datapoint $w_{0}$
in our minibatch we sample $t\sim Unif\{[0,T\}$, our corrupted sequence
$w_{t}\sim q(x_{t}|x_{0},t)$, and then calculate an average of the
cross entropy loss over the masked tokens. 

\subsection*{MLM Discrete Diffusion for Sudoku}

To make the application of MLM discrete diffusion to sudoku concrete,
we consider a dataset of sudoku solutions which we flatten row-wise
into vectors $\{w_{0}\in\mathbb{Z}^{81}\}$ containing tokens of digits
one through nine. Then, using the MLM cross entropy objective above,
we train our diffusion model, defined by a learned denoiser $\tilde{p}_{\theta}(\tilde{w_{0}}|w_{t})$.
Once our denoiser is learned, we can kick start the sudoku solution
generation process by first drawing an initial sample $w_{T}$ from
the absorbing state prior (which is simply all {[}MASK{]} tokens)
and then iteratively sample from the reverse transition probability
\[
p_{\theta}(w_{t-1}|w_{t})=\sum_{\tilde{w_{0}}}p(w_{t-1}|w_{t,}\tilde{w_{0}})p_{\theta}(\tilde{w_{0}|}w_{t})
\]

for each timestep $t\in[T,1]$ to generate our datapoint $\hat{w}_{0}.$ 

This above process takes a completely informationless sequence of
$[MASK]$ tokens and generates a (hopefully realistic and valid) sudoku
solution; however, it is simply \emph{a }sudoku solution. To generate
solutions corresponding to some initial sudoku board $x$, we use
infilling generation. At each time step $t\in[T,0]$ in the reverse
process (including the prior sample), given some partially filled
initial board $x$ with non-empty cells/tokens with indices $i\in I$
, we replace all tokens in the diffusion output $w_{t}$ with indices
$i\in I$ to be the corresponding token in the initial board $x$,
or $\forall i\in I\quad[w_{t}]_{i}:=[x]_{i}$. This ensures that our
reverse generation process generates a sudoku solution conditioned
on the given initial sudoku board. 

\subsection*{Guiding MLM Discrete Diffusion Models}

Above we saw that we can train a MLM discrete diffusion model to learn
the distribution of sudoku puzzles and also condition on a given initial
board to generate a corresponding solution. 

While this is reasonably performant, we can improve the output generation
process by incorporating guidance from some relevant value function
. In our case, we want to ensure that the solution satisfy the constraints
of sudoku (no digit must be duplicated in any given row, column, or
block) so we can incorporate some value function $v(w)$ which scores
how well a proposed solution satisfies the constraints. To get this
value function in practice, given a dataset of some (partially incorrect)
sudoku solutions and the number of row/column/block constraints violated,
we can train a network in a supervised manner to predict how few constraints
a given proposed sudoku solution violates. 

The immediate challenge with incorporating a value function $v(w)$
is that the value function might have a discrete domain: in our case,
the value function scores a sudoku board $w$ which is a sequence
of discrete tokens. A discrete domain is not differentiable, which
poses a question of how the value function signal would be incorporated
in the diffusion generation process. 

One workaround is to shift to a continuous domain rather than a discrete
domain. For example one could embed all sudoku boards $w$ not as
discrete tokens but as real valued sequences $w\in\mathbb{R}^{81}$.
Alternately one could train a seperate value function that is defined
not on discrete sequences $w$ but rather some continuous hidden logits
$h$ corresponding to the discrete sequences. This is in fact what
\cite{gruver2023protein} do with protein sequences: for any given (noisy)
sequence $w_{t}$ in the reverse diffusion process they take hidden
states (which hidden layer is left as a choice parameter) $h_{t}$
from a trained diffusion decoder $\tilde{p}_{\theta}(\tilde{w_{0}}|w_{t})$
and then utilize a value function defined on the (noisy) hidden states
$v(h_{t})$ . Since by construction $v$ is differentiable with continuous
inputs $h,$ we can take gradients of $v(h)$ to guide the original
logits $h_{t}$ from the diffusion denoiser $\tilde{p}_{\theta}(\tilde{w_{0}}|w_{t})$.
However, this requires training a value function based on noisy hidden
states and moroever requires the value function to be tied down to
the outputs of one specific trained diffusion model $\tilde{p}_{\theta}(\tilde{w}_{0}|w_{t})$. 

In many settings, this assumption may not be applicable and we may
only have access to a value function that is defined on the actual
outputs: say a function $v(w)$ that scores how constraint satisfying
a given discrete sudoku board $w$ is or as a more practical example
a function $f(p)$ which scores how correct a program $p$ is. 

To work around this, we can rely on the gumbel softmax trick, explained
briefly below. 

\subsubsection*{Gumbel Softmax Trick }

The Gumbel Softmax trick introduced by \cite{jang2017categorical} allows for
gradients to flow through a discrete or categorical variable.

\begin{figure}
\begin{centering}
\includegraphics[scale=0.3]{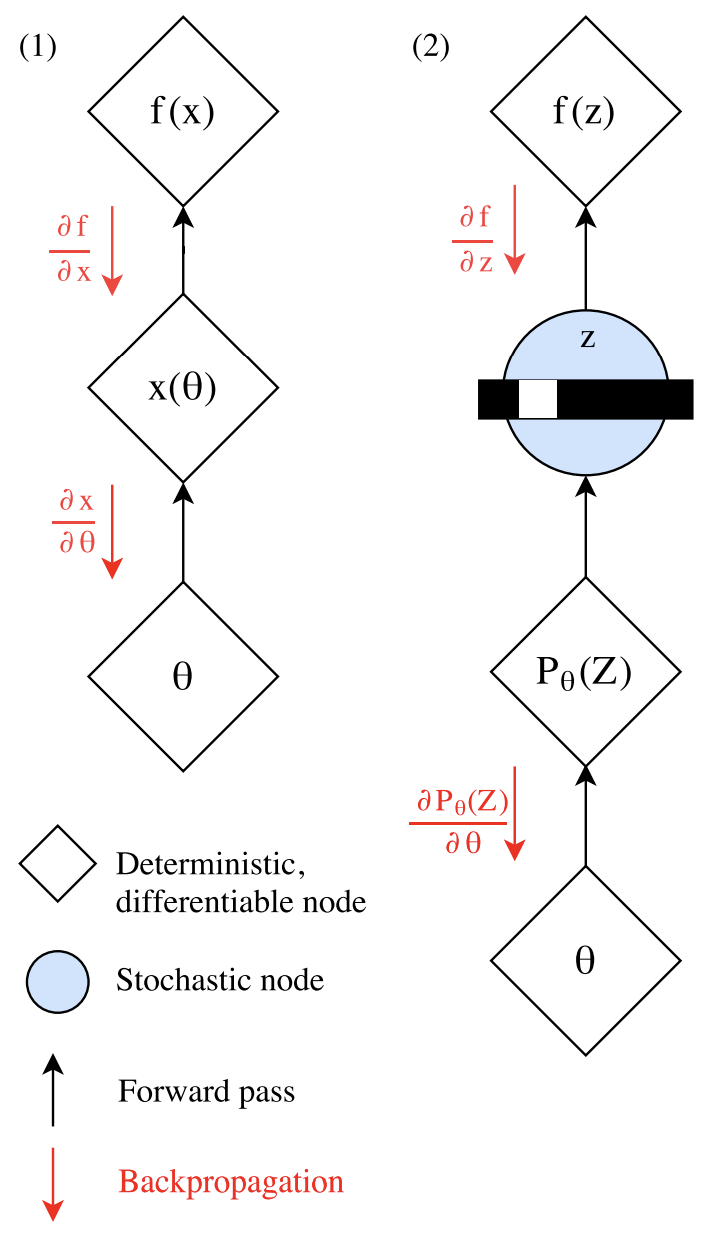}\caption{Gradient Flow With Categorical Variable}
\par\end{centering}
\end{figure}

In our case we want the gradients to flow with our setup $h\to w\to v(w)$;
$w\sim Cat(\pi(h))$ is sampled from the categorical probability distribution
$\pi$ defined by logits $h$ and then our value function $v$ scores
the discrete sample $w$. The problem is that we have a discrete categorical
sample $w$ in the middle, which prevents calculation of $\nabla_{h}w$
(and for that matter $\nabla_{\pi}w)$ without any tricks. For simplicity
let us now imagine that our sequence has length one: this means that
our ``sequence'' is a categorical variable sampled from some probability
distribution vector $\pi$ with class probabilities $\pi_{i}$. (In
reality, with a sequence of length $L,$ we would need a vector of
class probabilities for each token in the sequence.) 

More than half a century ago Gumbel proposed an efficient trick called
the ``Gumbel trick'' to draw samples $w$ from a categorical distribution
defined by a vector of class probabilitites $\pi$: 
\[
w=onehot(argmax_{i}[g_{i}+log\pi_{i}])
\]
where $g_{i}$ are i.i.d $Gumbel(0,1)$ variables. 

\cite{jang2017categorical} introduces a continuous approximation of the argmax
using softmax, explaining the name ``Gumbel softmax trick''. As
$\tau\to0,$ the vector $y$ defined by 
\[
y_{i}=\frac{exp((log\pi_{i}+g_{i})/\tau)}{\sum_{j}exp((log\pi_{j}+g_{j})/\tau)}
\]
 approaches the one-hot vector $w$ defined by the argmax. This is
a continuous relaxation and we see that our vector $y$ ends up differentiable
with respect to the probabilities $\pi$, albeit being a continous
vector rather than a discrete vector. With this approximation we now
have a work around to approximate the gradient of our sample $\nabla_{\pi}w$
using the approximation $\nabla_{\pi}y$. 

However, we do not need to keep our output $y$ continuous. Similar
to the straight through estimator, for the forward pass we can discretize
our continuous $y$ using argmax, leading to a desired discrete categorical
variable $w$ that can be passed to $v(w)$; for the backpropagation
step we can swap out the intractable $\nabla_{\pi}w$ for our approximation
$\nabla_{\pi}y$. 

\subsubsection*{Adding guidance in reverse process}

Now that we can flow gradients from our constraint value function
$v(w)$ defined on sudoku board sequences$,$ we can incorporate such
gradients in our reverse sample generation process. 

We do this by working in the logit space and guiding the initial logits
from the diffusion denoiser $\tilde{p}_{\theta}(\tilde{w_{0}}|w_{t})$.
Naively, one might just simply add the gradient directly, such as
$h'=h+\nabla_{h}v(gum(h))$. However this poses a tradeoff between
generating realistic samples faithful to the data distribution (coming
from diffusion logits $h$) and generating samples which maximize
the constraint value function $v(w)$. 

\cite{dathathri2020plug} in their work on plug and play language models
handle this by regularizing the guided logits to the initial logits.
Borrowing from this, our entire guided update step looks like
\[
h^{i+1}=h^{i}+\nabla_{h}v(gum(h^{i}))+\lambda\nabla_{h}KL(\pi(h^{0})||\pi(h^{i}))
\]
 where $i$ represents the update index and $h^{0}$ are the initial
logits. 

Now at each reverse sampling timestep we update our logits with multiple
regularized gradient ascent update steps. This leads the reverse process
to bias towards final samples $w_{0}$ that end up having a high constraint
satisfaction score, leading to better solutions.

The entire sampling algorithm is shown in Figure 3.

\begin{figure}

\begin{centering}
\includegraphics[scale=0.2]{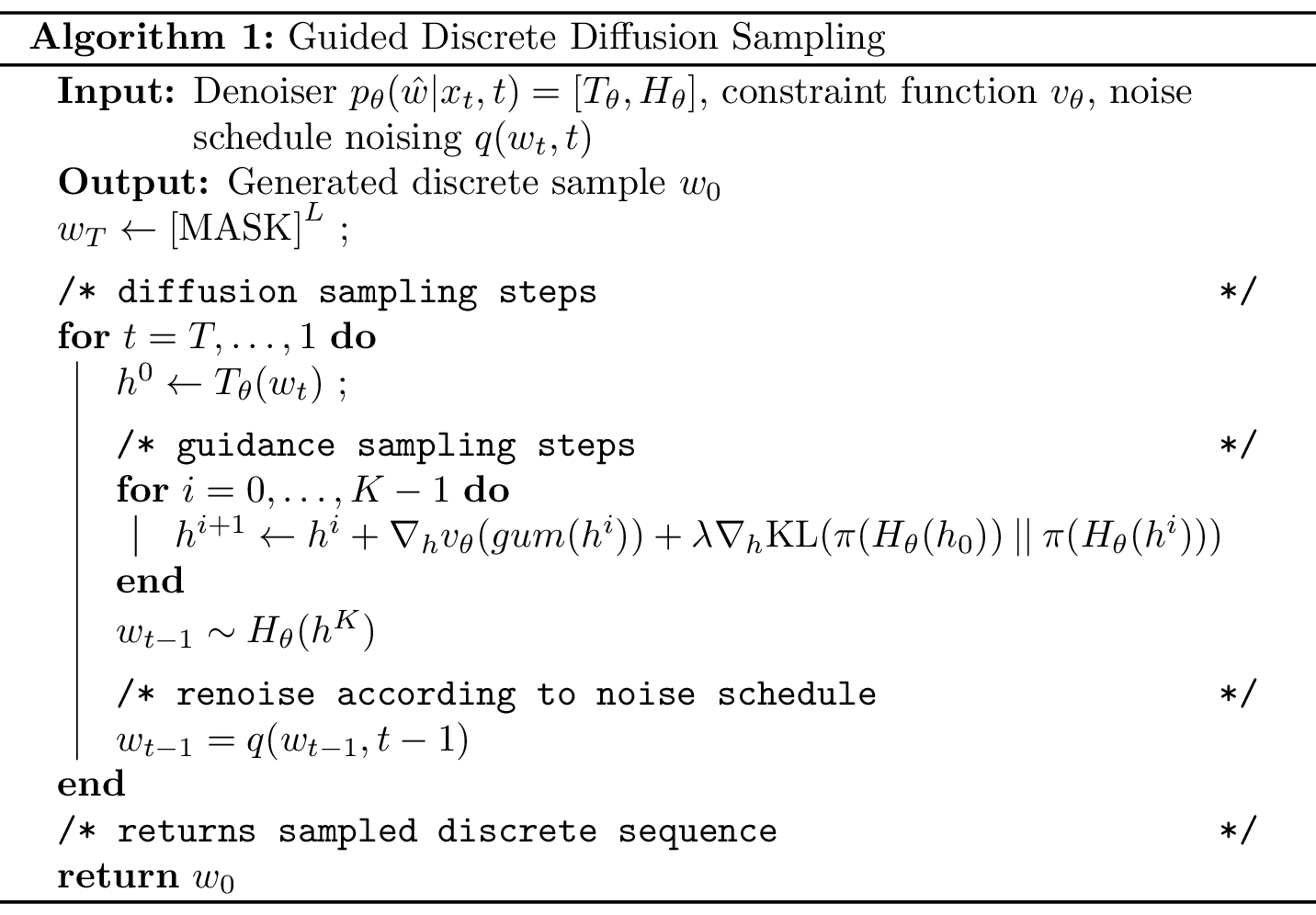}\caption{Guided MLM Sampling Algorithm}
\par\end{centering}
\end{figure}

The impact of guidance on solve rate is shown in the results, Figure
4. We see that by adding constraint guidance we are able to increase
the solve rate from 85.2\% to 90.6\%.

\subsection*{Score Entropy Discrete Diffusion}

Score Entropy Discrete Diffusion (SEDD) introduced by \cite{lou2023discrete} is a competitive, state of the art discrete diffusion model
which outperforms similarly sized GPT models on perplexity scores.
(An apt treatment on score entropy discrete diffusion models is out
of the scope of this post, but for a introduction Lou's blog post
is a good starting point (Lou 2024)).

In comparison to previous diffusion models such as D3PM or the canonical
DDPM, score entropy discrete diffusion is based on a \emph{continuous
time} markov chain. 

Now, for some categorical variable $x\in[K]$ with class probability
mass vectors $p\in\mathbb{R}^{K}$, we can characterize the probability
distribution $p$ evolving over continuous time, $\{p_{t}\},$ using
a differential equation rather than a bunch of discrete transition
probabilities:
\[
\frac{dp_{t}}{dt}=Q_{t}p_{t}\qquad p_{0}\approx p_{data}
\]

Here $Q_{t}\in\mathbb{R}^{N\times N}$ is our diffusion matrix which
evolve our probability vectors and define our forward markov chain.
(Similar to our explanation of D3PM, we consider the case of only
a single categorical variable: we will see why this is sufficient
even when working with discrete sequences with many tokens later on.)

A neat result shows that with our forward continuous time markov chain
given by $Q_{t}$ there exists a corresponding reversal continuous
time markov chain given by a reversal diffusion matrix $\bar{Q_{t}}$:
\[
\frac{dp_{T-t}}{dt}=\bar{Q}_{T-t}p_{T-t}
\]
 where we can define our reversal matrix $\bar{Q_{t}}$ in terms of
the forward matrix $Q_{t}$: $\bar{Q}_{t}(y,x)=\frac{p_{t}(y)}{p_{t}(x)}Q_{t}(x,y)$
and $\bar{Q}_{t}(x,x)=-\sum_{y\ne x}\bar{Q_{t}}(y,x)$.

In practice, we need to simulate the forward or reverse process and
so for computational feasibility we use an approximation to this differential
equation; in particular we consider a first order or linear Euler
approximation and we have our approximate transition probabilities
as
\[
p(x_{t+\Delta t}=y|x_{t}=x)\approx\delta_{xy}+Q_{t}(y,x)\Delta t
\]

and 
\[
p(x_{t-\Delta t}=y|x_{t}=x)\approx\delta_{yx}+\bar{Q}_{t}(y,x)\Delta t=\delta_{yx}+\frac{p_{t}(y)}{p_{t}(x)}Q_{t}(x,y)\Delta t
\]

where both approximations are accurate up to error $O(\Delta t^{2})$
and $\delta$ represents the dirac-delta function. 

Now given a defined forward process $Q_{t}$, if we want to generate
samples using the reverse process we can simply look at the reverse
transition probability above and see that all we need is the ratio
$\frac{p_{t}(y)}{p_{t}(x)}$. At a high level what SEDD is doing is
learning this ratio using a deep network, $s_{\theta}(y)_{x}=\frac{p_{t}(y)}{p_{t}(x)}$,
and then using this learned network in the reverse process simulation
to generate samples. 

Now in reality, we are dealing not with a single categorical variable
but rather sequences $\mathbf{x}\in\mathbb{Z}^{L}$ of many categorical
variables. This significantly increases the number of all possible
ratios $s_{\theta}(\mathbf{y})_{\mathbf{x}}$ (in fact to exponential
complexity). Thus for computational tractability, following \cite{campbell2022continuous} they factorize the sequence and just have each token
or categorical variable evolve independently of each other. Now, given
that we are in a continuous time setting, the probability of two or
more variables in a sequence evolving at same point in time $t$ is
zero: thus we only keep track of one token in the sequence transitioning
for any time $t$. 

Notationally, with a sequence $\mathbf{x}=(x^{1}...x^{L})$, we only
need to keep track of some token with index $i$ transitioning to
some value $\hat{x}_{i}$ and so our network only has to learn the
ratio that differs by one token: $s_{\theta}(\mathbf{x},t)_{i,\hat{x_{i}}}\approx\frac{p_{t}(x^{1}...\hat{x^{i}}...x^{L})}{p_{t}(x^{1}...x^{i}...x^{L})}$. 

However, if we only change one token at each timestep, this means
that for any given evolution most tokens will remain the same doing
nothing and for long sequences $L\gg1$ our reverse process simulation
will take a very long time to generate a final sequence sample. As
a computational speed up, SEDD can employ tau-leaping, which is an
approximation method to speed up sampling by essentially independently
sampling each token in the sequence for each timestep. 

With these tricks, SEDD is able to generate samples of high cardinality
and dimensionality such as language sequences with reasonable computation. 

\paragraph*{SEDD for sudoku}

We train SEDD on sudoku board sequences $w$ represented as sequences
of categorical tokens as before. Given a trained SEDD network, to
solve a given initial sudoku board, we again employ conditional infilling:
in each timestep of the reverse sampling process, we replace the diffusion
output to contain the non-empty initial sudoku tokens. 

What we see in our results is that SEDD demonstrates state of the
art sample efficiency on SATNet, a common Sudoku benchmark. Other
models such as transformer or graph based supervised networks trained
on SATNet also achieve 100\% solve rate accuracy on SATNet and so
SEDD is not any more competitive than typical supervised methods.
However, it is notable that even with an order of magnitude less training
data, learning from only hundreds of puzzle solutions (no supervised
data needed) we get performant results.

\begin{figure}
\begin{centering}
\includegraphics[scale=0.4]{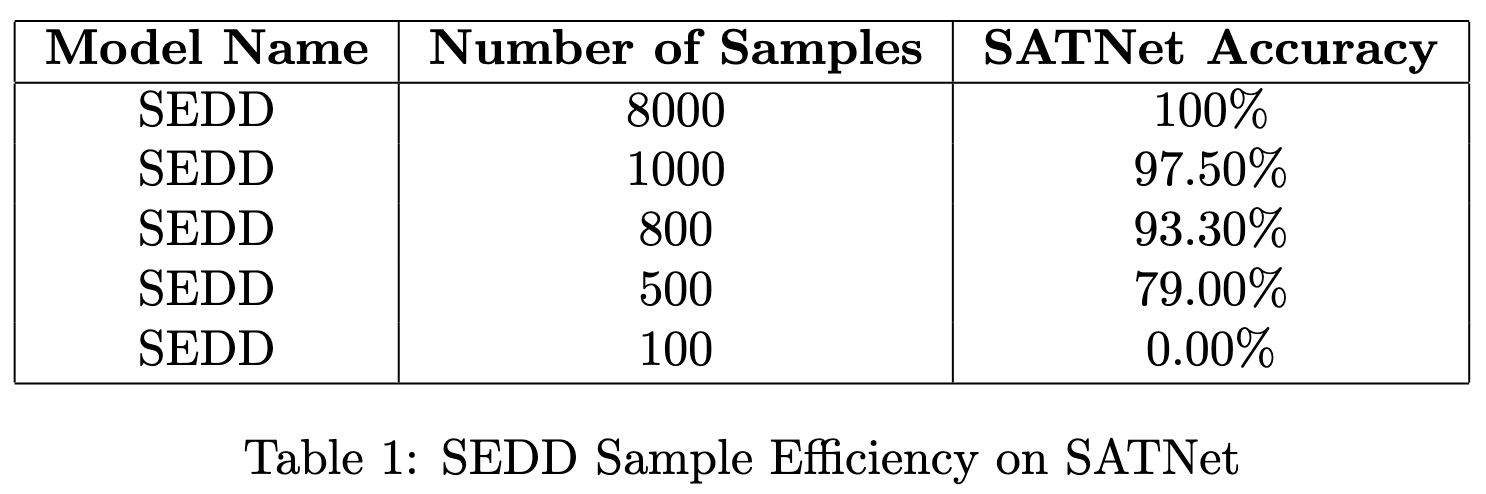}
\par\end{centering}
\end{figure}

\subsection*{Results}

\begin{figure}

\centering{}\includegraphics[scale=0.3]{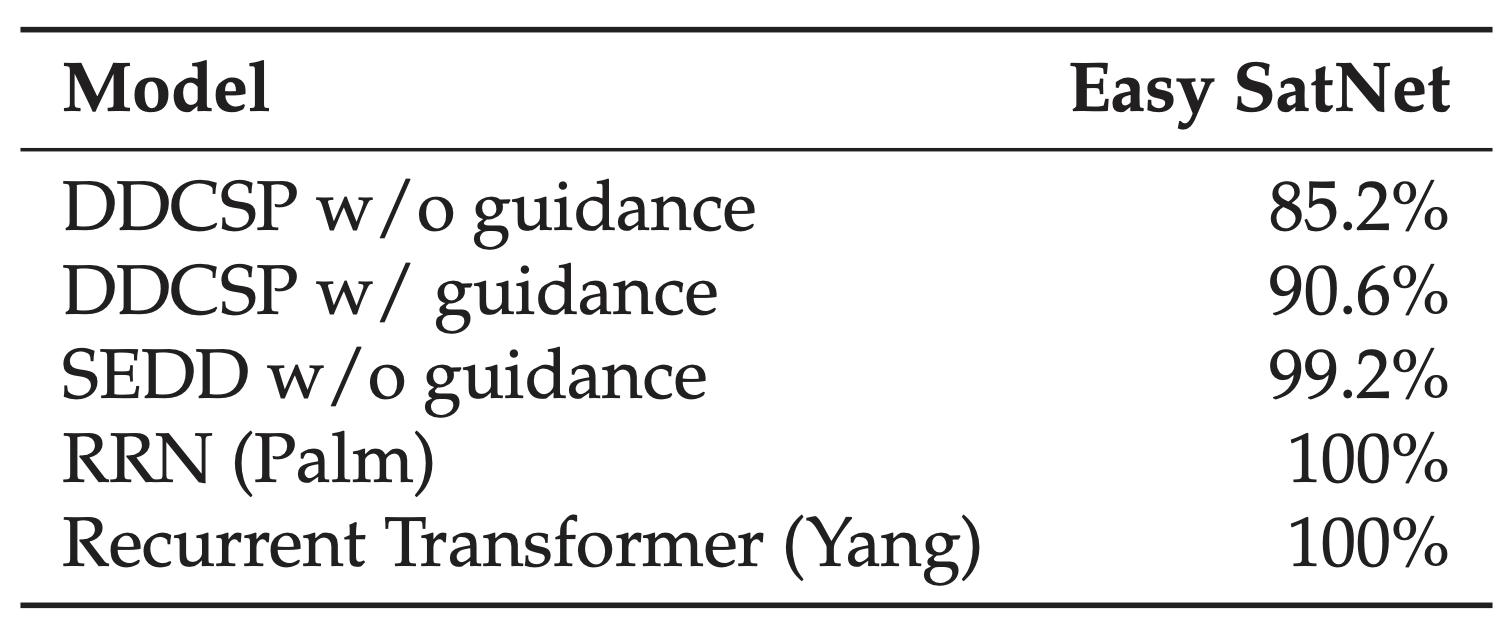}\caption{Sudoku Performance}
\end{figure}

We demonstrate the results of our discrete diffusion model against
some baselines. \cite{palm2018recurrent} is a graph neural network and \cite{yang2023learning} is a recurrent
transformer, both trained with supervised datasets. DDCSP is the MLM
discrete diffusion model. While DDCSP is not as performant as the
supervised networks, we see that with guidance performance increases
considerably. SEDD, the state of the art discrete diffusion model,
is competitive with the supervised baselines. 

\subsection*{Related Work}
\cite{cardei2025constraineddiscretediffusion} uses a similar guidance method and demonstrates performance on multiple tasks such as language and molecule generation. Our work was developed independently, with our work released in January, preceding the first draft of \cite{cardei2025constraineddiscretediffusion} in March.

\subsection*{Conclusion}

In summary, we have shown how discrete diffusion models offer competitive
performance on constraint satisfaction problems such as sudoku. By
employing a guidance technique, we are able to further improve diffusion
output solve rate. Moreover, we show that even with limited samples,
state of the art discrete diffusion models exhibit performant behavior
on sudoku benchmarks. A natural next step is to add guidance to the
score entropy discrete diffusion process and see how guidance improves
performance and sample efficiency. 

\subsection*{Acknowledgements}

I would like to thank Tim Hanson for reviewing this paper and providing
feedback. This work was made possible through the philanthropic support
of Schmidt Futures.

\bibliographystyle{plainnat} 
\bibliography{references} 

\begin{thebibliography}{10}
\providecommand{\natexlab}[1]{#1}
\providecommand{\url}[1]{\texttt{#1}}
\expandafter\ifx\csname urlstyle\endcsname\relax
  \providecommand{\doi}[1]{doi: #1}\else
  \providecommand{\doi}{doi: \begingroup \urlstyle{rm}\Url}\fi

\bibitem[Austin et~al.(2021)Austin, Johnson, Ho, Tarlow, and van~den
  Berg]{austin2021structured}
Jacob Austin, Daniel~D Johnson, Jonathan Ho, Daniel Tarlow, and Rianne van~den
  Berg.
\newblock Structured denoising diffusion models in discrete state-spaces.
\newblock \emph{Advances in Neural Information Processing Systems},
  34:\penalty0 17981--17993, 2021.

\bibitem[Campbell et~al.(2022)Campbell, Benton, De~Bortoli, Rainforth,
  Deligiannidis, and Doucet]{campbell2022continuous}
Andrew Campbell, Joe Benton, Valentin De~Bortoli, Thomas Rainforth, George
  Deligiannidis, and Arnaud Doucet.
\newblock A continuous time framework for discrete denoising models.
\newblock \emph{Advances in Neural Information Processing Systems},
  35:\penalty0 28266--28279, 2022.

\bibitem[Cardei et~al.(2025)Cardei, Christopher, Hartvigsen, Kailkhura, and
  Fioretto]{cardei2025constraineddiscretediffusion}
Michael Cardei, Jacob~K Christopher, Thomas Hartvigsen, Bhavya Kailkhura, and
  Ferdinando Fioretto.
\newblock Constrained discrete diffusion, 2025.
\newblock URL \url{https://arxiv.org/abs/2503.09790}.

\bibitem[Dathathri et~al.(2020)Dathathri, Madotto, Lan, Hung, Frank, Molino,
  Yosinski, and Liu]{dathathri2020plug}
Sumanth Dathathri, Andrea Madotto, Janice Lan, Jane Hung, Eric Frank, Piero
  Molino, Jason Yosinski, and Rosanne Liu.
\newblock Plug and play language models: A simple approach to controlled text
  generation.
\newblock \emph{arXiv preprint arXiv:1912.02164}, 2020.

\bibitem[Gruver et~al.(2023)Gruver, Stanton, Frey, Rudber, Isber, Fromer, and
  Wilson]{gruver2023protein}
Nate Gruver, Samuel Stanton, Nathan Frey, Tim G.~J. Rudber, Romann Isber,
  Jenna~Colleen Fromer, and Andrew~Gordon Wilson.
\newblock Protein design with guided discrete diffusion.
\newblock \emph{Advances in Neural Information Processing Systems}, 36, 2023.

\bibitem[Jang et~al.(2017)Jang, Gu, and Poole]{jang2017categorical}
Eric Jang, Shixiang Gu, and Ben Poole.
\newblock Categorical reparameterization with gumbel-softmax.
\newblock \emph{arXiv preprint arXiv:1611.01144}, 2017.

\bibitem[Lou et~al.(2023)Lou, Meng, and Ermon]{lou2023discrete}
Aaron Lou, Chenlin Meng, and Stefano Ermon.
\newblock Discrete diffusion modeling by estimating the ratios of the data
  distribution.
\newblock \emph{arXiv preprint arXiv:2310.16834}, 2023.

\bibitem[Luo(2022)]{luo2022understanding}
Calvin Luo.
\newblock Understanding diffusion models: A unified perspective.
\newblock \emph{arXiv preprint arXiv:2208.11970}, 2022.

\bibitem[Palm et~al.(2018)Palm, Paquet, and Winther]{palm2018recurrent}
Rasmus~Berg Palm, Ulrich Paquet, and Ole Winther.
\newblock Recurrent relational networks.
\newblock \emph{Advances in Neural Information Processing Systems}, 31, 2018.

\bibitem[Yang et~al.(2023)Yang, Ishay, and Lee]{yang2023learning}
Zhun Yang, Adam Ishay, and Joohyung Lee.
\newblock Learning to solve constraint satisfaction problems with recurrent
  transformer.
\newblock \emph{arXiv preprint arXiv:2110.13397}, 2023.

\end{thebibliography}

\end{document}